# Flow to Rare Events: An Application of Normalizing Flow in Temporal Importance Sampling for Automated Vehicle Validation*

Yichun Ye
The Key Laboratory of Road and
Traffic Engineering, Ministry of
Education
Tongji Universtiy
Shanghai, China
2331787@tongji.edu.cn

He Zhang
The Key Laboratory of Road and
Traffic Engineering, Ministry of
Education
Tongji Universtiy
Shanghai, China
zhanghe_tj@tongji.edu.cn

Ye Tian
The Key Laboratory of Road and
Traffic Engineering, Ministry of
Education
Tongji Universtiy
Shanghai, China
tianye@tongji.edu.cn

Jian Sun
The Key Laboratory of Road and
Traffic Engineering, Ministry of
Education
Tongji Universtiy
Shanghai, China
sunjian@tongji.edu.cn

Karl Meinke
Computer Science Institute, School of
Electrical Engineering and Computer
Science
KTH Royal Institute of Technology
Stockholm, Sweden
karlm@kth.se

*Abstract*—Automated Vehicle (AV) validation based on simulated testing requires unbiased evaluation and high efficiency. One effective solution is to increase the exposure to risky rare events while reweighting the probability measure. However, characterizing the distribution of risky events is particularly challenging due to the paucity of samples and the temporality of continuous scenario variables. To solve it, we devise a method to represent, generate, and reweight the distribution of risky rare events. We decompose the temporal evolution of continuous variables into distribution components based on conditional probability. By introducing the Risk Indicator Function, the distribution of risky rare events is theoretically precipitated out of naturalistic driving distribution. This targeted distribution is practically generated via Normalizing Flow, which achieves exact and tractable probability evaluation of intricate distribution. The rare event distribution is then demonstrated as the advantageous Importance Sampling distribution. We also promote the technique of temporal Importance Sampling. The combined method, named as TrimFlow, is executed to estimate the collision rate of Car-following scenarios as a tentative practice. The results showed that sampling background vehicle maneuvers from rare event distribution could evolve testing scenarios to hazardous states. TrimFlow reduced 86.1% of tests compared to generating testing scenarios according to their exposure in the naturalistic driving environment. In addition, the TrimFlow method is not limited to one specific type of functional scenario.

*Keywords—Rare Events, Temporal Distribution, Normalizing Flow, Importance Sampling, Accelerated Validation*

I. INTRODUCTION

Automated Vehicles (AVs), especially Highly Automated Vehicles (HAVs), aim to enhance the safety of the transportation system [1]. Despite significant advancements in autonomous driving technology, AV-related accidents inevitably lead to public distrust and suspicion [2, 3]. From the perspective of AV technology, there is a necessity for diagnosing and ameliorating its deficiencies. Regarding practical deployment, safety validation should also be complementarily promoted. It is essential to qualify and demonstrate the safety performance of AV in the naturalistic driving environment.

To avoid physical damage and extremely lengthy testing process, simulated testing has been widely received in recent years. Compared to road testing and closed-course testing, it allows us to give priority to generating challenging and unusual scenarios.

Despite the use of simulation, two basic issues still need to be addressed. Firstly, to reflect the safety performance (e.g., risk rate) of AV in the naturalistic driving environment, statistical quantification requires reliable naturalistic driving data. Secondly, because of "the Curse of Rarity" [4], an efficient validation process calls for higher exposure to risky rare events. To reconcile the two objectives, it is infeasible to generate testing scenarios according to their exposure in the naturalistic driving environment.

One workaround to this thorny problem is rare event simulation, which offers solutions for the computation of such rare event probability. Considering the characteristics of AV validation, there are several difficulties involved in the application of rare event simulation. AV validation is generally conducted within traffic scenarios. The temporal correlation of multiple background vehicles' states and maneuvers results in the high-dimensionality of the rare event distribution. The paucity of risky event samples exacerbates the difficulty of fitting high-dimensional distribution. In addition, during stepwise simulation, estimating the probability of continuous maneuver variables is far more challenging than that of discrete variables. As for the formulation of distributions, rare events can be irregularly distributed, which makes it hard to impose any conventional priori distribution formulation and to solve its parameters.

Existing research has provided us with valuable insights. To search for risky scenarios, Feng et al. [5] proposed a new criticality definition as a combination of exposure frequency and maneuver challenge. To describe the evolution of scenarios, in the latter research of Feng et al. [6], they

*Research supported by the National Key Research and Development Program of China [2021YFB2501202], and the National Natural Science Foundation of China under Grant number [52172391].

described the testing scenarios with the states of AV, the states of background vehicles, and their maneuvers in discrete forms. Meinke and Khosrowjerdi [7] presented a modeling language for testing scenario evolution. They proposed constrained active machine learning for dynamic analysis by inferring chains of intersecting automaton models. To estimate the probability of the rare events, our previous research [8] applied Masked Autoregressive Flow (MAF) [9], a type of Normalizing Flow (NF) model, to fit the distribution of collision scenarios. As a generative probabilistic model, NF enables exact and efficient inference of the probability of latent variables through a series of invertible mappings with tractable Jacobian [10]. To reweight the probability measure, Zhao et al. [11] and Huang et al. [12] have applied Importance Sampling [13] to obtain unbiased estimation of risk rate in traffic scenarios like Cut-in, Lane-changing and Car-following. Arief et al. [14] supposed that the relative states between background vehicles and AV followed Gaussian distribution and proposed Deep Importance Sampling to relocate its sampling center. By increasing the exposure to risky events, the required number of tests can be greatly reduced.

The above methods addressed the challenges of AV validation from multiple perspectives. However, existing methods either lack consideration of the temporality of scenario variables, or use pre-generated logs as the scenario input, losing the flexibility of stepwise interaction. The discretization of scenario variables also simplifies the practical problem. Additionally, if the distribution fitting of risky rare events heavily relies on the known risky samples, collecting sufficient samples will impose a strenuous burden. There is a lack of research on the general formulation of risky rare event distribution.

In this work, we achieve **t**emporal **Im**portance Sampling on **r**are events based on Normalizing **Flow** (named as **TrimFlow**). The core idea of TrimFlow is to represent, generate, and reweight the distribution of risky rare events as shown in Fig.1. The evolution of the testing scenario is assumed to follow the Markov property. By introducing the Risk Indicator Function at each timestep, the formulation of risky rare events distribution can be precipitated out of the naturalistic driving distribution. The component distributions of rare event distribution, given such formulation, are generated via NF. The rare event distribution is then demonstrated as the advantageous Importance Sampling distribution. Combined with the accept-reject sampling method, we also achieve the temporal Importance Sampling. As a tentative practice, we execute the TrimFlow to estimate the collision rate of our System Under Test (SUT) in Car-following scenarios. The contributions of TrimFlow are as follows:

- Promoting a general formulation for the temporal distribution of risky events and overcoming the limitation of only relying on rare event samples to fit its distribution;
- Retaining the continuity of maneuver variables in test scenarios and demonstrating the effectiveness of NF in probability evaluation;
- Achieving significant tests reduction compared to tests based on the naturalistic driving distribution;

The rest of this paper is organized as follows. In Section II, we review relevant research. Section III elaborates on the detailed design of the TrimFlow methodology. In Section IV, we present the implementation procedure of TrimFlow. In Section V, the proposed method is executed in Car-following scenarios. In Section VI, we summarize our findings and introduce plans for further research.

## II. RELATED WORKS

### A. Formulation of Testing Scenarios

As for the elements of testing scenarios, there exist brief reviews on scenario specification language in [15] and [16]. Bagschik et al. [17] presented an ontology-based, five-layer model for scenario representation. It included roads (L1), traffic infrastructure (L2), manipulation of L1 and L2 together (L3), objects (L4) and environment (L5).

The scope and form of testing scenarios are also diversely adopted. Some research only considered the initial inputs of scenario variables, which defined the scenarios as a trigger and a sequence of ruled-based processes [18]. Some research generated scenarios as fixed-length trajectories or maneuvers [19]. The above two forms of testing scenarios simplify the process of scenario evolution. To emphasize the interaction of the SUT with the background vehicles, determining the maneuvers of background vehicles step by step, i.e., stepwise generation of testing scenarios, is more reasonable and adaptable [6].

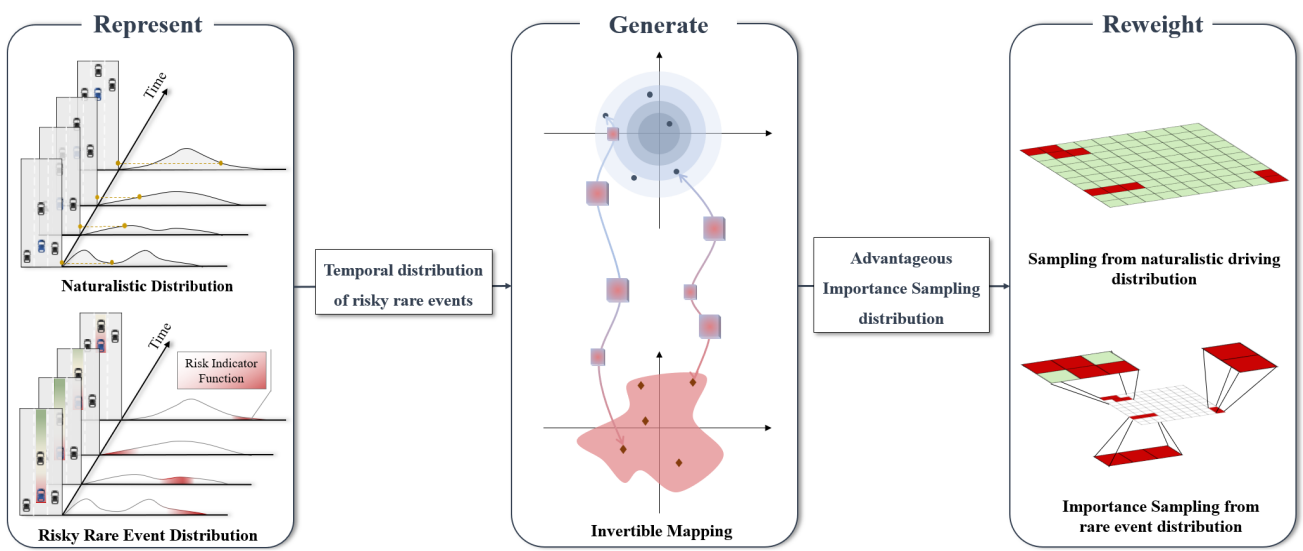

Fig. 1. Framework of TrimFlow. The AV validation is closely related to the evolution of scenario variables. By introducing the Risk Indicator Function, the temporal distribution of risky rare events is theoretically precipitated out of the naturalistic driving distribution. Its distribution components are practically generated via NF. The rare event distribution is then demonstrated as the advantageous Importance Sampling distribution. By reweighting its probability, we eventually get the unbiased estimaiton of risk rate.

*B. Normalizing Flow*

Different from other generative models such as Variational Autoencoder (VAE) and Generative Adversarial Network (GAN), NF can achieve explicit and tractable probability estimation. There exist various formations of NFs. Dinh et al. [20] proposed real-valued non-volume preserving (real NVP) transformations. They aimed to deal with high-dimensional and highly structured datasets. In the research of Kingma et al. [21], they developed the Inverse Autoregressive Flow (IAF) to achieve estimation accuracy, which is also suited to high-dimensional tensor variables. Papamakarios et al. [9] further improved the efficiency of the estimation process. They presented Masked Autoregressive Flow (MAF) using the Masked Autoencoder for Distribution Estimation (MADE), which can be trained on parallel computing architectures.

There are some trials on temporal NFs [22, 23]. However, they are limited to learn the features of stochastic dynamics so far.

*C. Acclerated Safety Validation*

Safety validation is primarily concerned with the failure probability of the SUT [24]. Zhao et al. [11] first applied Importance Sampling on accelerated evaluation of the risk rate of AVs. It is a classical variance reduction technique used to approximate an equivalent expectation with the Monte Carlo method, while sampling from an adjusted distribution where rare events are more likely to occur. Zhang et al. [25] proposed surrogate-based Monte Carlo method to reduce the massive tests required by the Law of Large Numbers. Feng et al. [6] increased the frequency of adversarial maneuvers of background vehicles through a dense deep-reinforcement-learning approach. Fu et al. [26] applied Subset Simulation to efficiently approach the failure zone represented by collision scenarios.

It can be concluded that accelerated safety validation aims at computing the probability of rare events with less tests. The selection of testing scenarios cannot solely refer to naive sampling methods, and the distribution pattern of scenarios needs to be adjusted.

III. METHODOLOGY

In this section, we present the detailed methodology of TrimFlow. Specifically, we formulate the temporal distribution of naturalistic driving scenarios and the ideal representation of risky rare event probability. We consider using the Importance Sampling method for unbiased estimation of the risk rate. Due to the practical infeasibility of the ideal representation, refined Risky Indicator Function is introduced to obtain the learnable Importance Sampling distribution. As an intricate distribution that cannot be parameterized using classical distribution forms, probability density inference is achieved by NF. In the following description, "scenario" refers to the time sequence of one testing case, while "scene" refers to the scenario slice at one timestep.

*A. Ideal Representation of Risky Event Probability*

A naturalistic driving scenario can be described as the serial connection process of each timestep scene. The transition equation is defined as
$$s_{t+1} = C(s_t, m_t), \quad (1)$$
where $s_t$ denotes the states of AV and background vehicles at $t$ timestep, and $m_t$ is the maneuvers of background vehicles at $t$ timestep. The transition $C$ refers to the formulation of kinematics in the simulation. The $s_{t+1}$ of AV and all background vehicles can be completely updated based on $s_t$ and $m_t$. Given the first $s_1$, Equation (1) is equivalent to
$$s_{t+1} = C(s_1, m_1, \dots m_t). \quad (2)$$
Based on the above definition, a testing scenario with a temporal length of $T$ can be decomposed into
$$X_T = [s_1, m_1, m_2, \dots m_T]. \quad (3)$$
Its probability is
$$p(X_T) = p(s_1, m_1, m_2, \dots m_T). \quad (4)$$
The maneuvers $m_t$ performed by background vehicles are temporally correlated with $s_t$. Specifically, the probability of $m_t$ is the conditional probability of $s_t$ as
$$p(m_t|s_t) = p(m_t|s_1, m_1, \dots m_{t-1}). \quad (5)$$
Given that the maneuver-decision process of background vehicles follows the Markov property, we have
$$p(X_T) = p(s_1) \prod_{t=1}^{T} p(m_t|s_t). \quad (6)$$
The probability expression of testing scenarios is consistent with [6]. This expression is not limited to one specific type of functional scenario. However, learning the conditional probability density function in (6) still presents significant challenges, especially for continuous maneuver variables and multiple background vehicles. According to the definition of conditional probability, we further simplify (6) as
$$p(X_T) = p_s(s_1) \prod_{t=1}^{T} \frac{p_{m,s}(m_t, s_t)}{p_s(s_t)}, \quad (7)$$
where $p_{m,s}$ and $p_s$ respectively represent the joint probability density function of $m_t, s_t$ and $s_t$ at each scene following the naturalistic driving distribution.

Risk rate estimation of risky events $\varepsilon$ needs an Indicator Function as
$$I_\varepsilon(X_T) = \begin{cases} 1, X_T \in \varepsilon \\ 0, X_T \notin \varepsilon \end{cases}. \quad (8)$$
It is noted that the uncertainty of testing results is not considered in (8). By introducing the Indicator Function, the risk rate $\mathbb{P}(\varepsilon)$ of AV can be formulated as the expected value of $I_\varepsilon(X_T)$. We have
$$\mathbb{P}(\varepsilon) = \mathbb{E}_{X_T \sim p(X_T)}[I_\varepsilon(X_T)] = \int p(X_T) I_\varepsilon(X_T) \, dX_T. \quad (9)$$
However, it is impractical to obtain $\mathbb{P}(\varepsilon)$ through integration. According to the Law of Large Numbers, an estimated value $\widehat{\mathbb{P}}(\varepsilon)$ can be calculated by generating $n$ independent samples as
$$\widehat{\mathbb{P}}(\varepsilon) = \frac{1}{n} \sum_{i=1}^{n} I_\varepsilon(X_T^i). \quad (10)$$
The accuracy of (10) is reflected by the relative half width $\omega$ of the confidence interval. When the confidence level is $100 \times (1-\beta)\%$, we need to sample at least $n$ times to guarantee that $\omega$ is not greater than the threshold $b$. As shown in (10), the required $n$ should be
$$n \geq \frac{1-\mathbb{P}(\varepsilon)}{\mathbb{P}(\varepsilon)} \cdot \frac{z_{\beta/2}^2}{b^2}, \quad (11)$$

where $z_{\beta/2} = \Phi^{-1}\left(1 - \frac{\beta}{2}\right)$ and $\Phi^{-1}$ represents the inverse cumulative distribution function of the normal distribution $N(0,1)$.

According to the technique of Importance Sampling, a new distribution $q(X_T)$ is introduced to raise the exposure of risky rare events. We have

$$\mathbb{P}(\varepsilon) = \int \frac{p(X_T)}{q(X_T)} q(X_T) I_\varepsilon(X_T) dX_T$$
$$= \mathbb{E}_{X_T \sim q(X_T)}\left[\frac{p(X_T)}{q(X_T)} I_\varepsilon(X_T)\right], \quad (12)$$
$$\widehat{\mathbb{P}}(\varepsilon) = \frac{1}{n}\sum_{i=1}^n \frac{p(X_T^i)}{q(X_T^i)} I_\varepsilon(X_T^i). \quad (13)$$

The efficiency of Importance Sampling is reflected by the variance. Minimizing variance is equivalent to

$$\min_q \mathbb{E}_{X_T \sim q(X_T)}\left[\frac{p^2(X_T)}{q^2(X_T)} I_\varepsilon^2(X_T)\right], \quad (14)$$

The argument $q$ here is a probability density function. So, this is a problem of functional analysis. Define the functional $J[q]$, we have

$$J[q] = \int \frac{p^2(X_T)}{q(X_T)} I^2(X_T) dX_T, \quad (15)$$

Set the variation equal to zero to find extremum, we have

$$\delta J = -\int p^2(X_T) I^2(X_T) \frac{1}{q^2(X_T)} \delta q(X_T) dX_T = 0. \quad (16)$$

Optimal solution $q(X_T)$ can be solved by the calculus of variations as

$$q(X_T) = c\, I_\varepsilon(X_T)\, p(X_T), \quad (17)$$

where $c$ is a constant. Equation (17) provides an ideal representation of risky event probability $q(X_T)$.

### B. Learnable Importance Sampling Distribution

Although the Indicator Function $I_\varepsilon(X_T)$ can precipitate the risky event distribution out of the naturalistic driving distribution, it has obvious limitation for distribution fitting. Because of the rarity of $\varepsilon$, $q(X_T)$ will be an extremely sparse and discrete distribution.

The essence of Importance Sampling ensures that the estimated $\widehat{\mathbb{P}}(\varepsilon)$ is theoretically unbiased regardless of which type of Importance Sampling distribution is applied, with only a difference in the saving of the required $n$. Therefore, the workaround is to replace $I_\varepsilon(X_T)$ and devise a learnable Importance Sampling distribution $q^*(X_T)$ which is close to $q(X_T)$ and easier to fit. The new Indicator Function needs to have the following properties:

- Continuous range between 0 to 1;
- Increasing monotonically with the level of risk;
- The largest absolute value of the derivative at 1;
- Able to calculate at each timestep;

Based on the above requirements, Risky Indicator Function $D(X_T) = e^{-ttc(s_t)}$ during the testing process is chosen to replace $I_\varepsilon(X_T)$ in (17). So, we propose a form of Importance Sampling distribution $q^*(X_T)$ as

$$q^*(X_T) = c\, D(X_T)\, p(X_T). \quad (18)$$

Combined with (7) we have

$$q^*(X_T) = c\, D(X_T)\, p_s(s_1) \prod_{t=1}^T \frac{p_{m,s}(m_t, s_t)}{p_s(s_t)}. \quad (19)$$

It can also be equivalently expressed as

$$q^*(X_T) = \frac{D(X_T)}{D(s_1)} \lambda_1 p_s(s_1) \prod_{t=1}^T \frac{\lambda_t p_{m,s}(m_t, s_t)}{\lambda_t p_s(s_t)}, \quad (20)$$

when $\lambda_t = \frac{D(s_t)}{\int p_s(s_t) D(s_t) ds_t}$, $c = \frac{1}{\int p_s(s_1) D(s_1) ds_1}$.

Thus, we set

$$q^*_{m,s}(m_t, s_t) = \lambda_t p_{m,s}(m_t, s_t) = \frac{p_{m,s}(m_t, s_t) D(s_t)}{\int p_s(s_t) D(s_t) ds_t}, \quad (21)$$
$$q^*_s(s_t) = \lambda_t p_s(s_t) = \frac{p_s(s_t) D(s_t)}{\int p_s(s_t) D(s_t) ds_t}. \quad (22)$$

Omitting the coefficients that are not conducive to learning, we obtain

$$\mathbb{P}(\varepsilon) = \mathbb{E}_{X_T \sim q(X_T)}\left[\frac{p_s(s_1)}{q^*_s(s_1)} \prod_{t=1}^T \frac{p_{m,s}(m_t, s_t) q^*_s(s_t)}{q^*_{m,s}(m_t, s_t) p_s(s_t)} I_\varepsilon(X_T)\right], \quad (23)$$

where $q^*_{m,s}$ and $q^*_s$ respectively represent the joint probability density function of $m_t, s_t$ and $s_t$ at each scene following the distribution of risky rare events.

Finally, the estimated risky rate is

$$\widehat{\mathbb{P}}(\varepsilon) = \frac{1}{n}\sum_{i=1}^n \left(\frac{p_s(s_1^i)}{q^*_s(s_1^i)} \prod_{t=1}^T \frac{p_{m,s}(m_t^i, s_t^i) q^*_s(s_t^i)}{q^*_{m,s}(m_t^i, s_t^i) p_s(s_t^i)} I_\varepsilon(X_T^i)\right). \quad (24)$$

### C. Normalizing Flow for Probability Density Inference

Let $X_T$ denote a set of scenario variables and $Z_T$ denote a set of latent variables, where $X_T, Z_T \in \mathbb{R}^D$, $X_T \sim q_\theta(X_T)$, $Z_T \sim p_z(Z_T)$. The $p_z(Z_T)$ is a basic simple distribution, and $q_\theta(X_T)$ is the learned distribution which approximates the Importance Sampling $q^*(X_T)$. By applying a sequence of invertible transformations $F$, $Z_T$ is mapped as

$$Z_T = F(X_T), F = f_K \circ \ldots f_2 \circ f_1. \quad (25)$$

By applying multiple transformations, the transformation of variable probability is defined as

$$\log q_\theta(X_T) = \log p_z(Z_T) + \sum_{k=1}^K \log \left|det J_{f_k}\right|, \quad (26)$$

where $J_{f_k}$ denotes the Jacobian determinant of each transformation $J_{f_k}$ at $X_{T,k}$. A sample $Z_T$ is easily drawn in the latent space, and its inverse sample $F^{-1}(Z_T)$ generates a sample $X_T$ from $q_\theta(X_T)$. Computing the probability on $X_T$ is accomplished by (26).

To improve the efficiency of probability calculation, real NVP is adopted herein. The invertible transformation is accomplished by affine coupling layers. Given a $D$ dimensional input $X_T$ and $d < D$, the output $y$ of an affine coupling layer follows the equations:

$$y_{1:d} = X_{T_{1:d}}, \quad (27)$$
$$y_{d+1:D} = X_{T_{d+1:D}} \odot exp\left(\alpha(X_{T_{1:d}})\right) + \mu(X_{T_{1:d}}), \quad (28)$$

where the learned $\alpha$ and $\mu$ stand for scale and translation, and $\odot$ is the Hadamard product or element-wise product. The output $y$ of one affine coupling layer becomes the input of the next layer, ensuring the transformation continuity to latent variables $Z_T$. Fig.2 illustrates one variable transformation with one affine coupling layer.

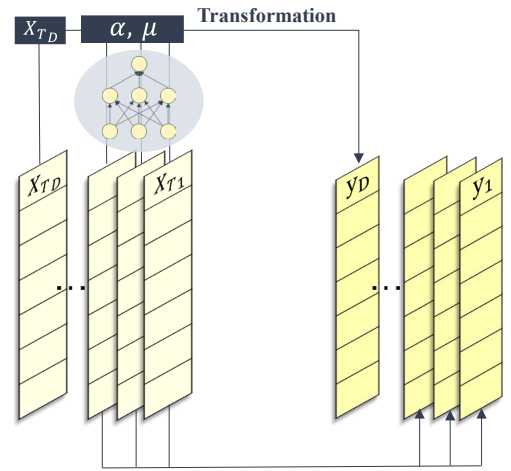

Fig. 2. $X_{T_D}$ transformation with one affine coupling layer

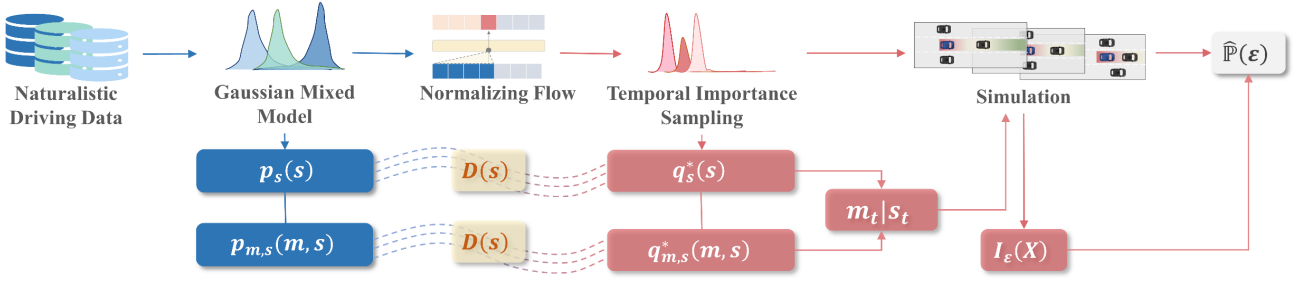

Fig. 3. Implementation Procedure for TrimFlow. Gaussian Mixed Model is used to fit $p_{m,s}$ and $p_s$ in the naturalistic driving environment. The $q^*_{m,s}$ and $q^*_s$ of risky rare events are then learnt as the fouction of $p_{m,s}$, $p_s$ and the Risk Indicator filter $D(s)$. Temporal Importance Sampling is adopted to sample the maneuvers of background vehicles and achieve unbiased estimation.

During the training process of Flow model, its general objective is to maximize $\log q_\theta(X_T)$ through the inverse mapping of the latent variables. The loss function is written as the negative log-likelihood function of $X_T$.

We consider using normal distribution as the basic simple distribution $p_z(Z_T)$. Since $q^*_{m,s}(m,s)$ and $q^*_s(s)$ will be learned separately, we take $q^*_s(s)$ as an example and the adopted loss function is derived as

$D_{KL}[q^*_s(s)\|q_{\theta,s}(s)]$

$= \int \frac{p_s(s)D(s)}{\int p_s(s)D(s)ds}[\log \frac{p_s(s)D(s)}{\int p_s(s)D(s)ds} - \log q_{\theta,s}(s)]ds.$  (29)

Combine (29) and (26) and we have

$D_{KL}$
$= \int \frac{c_1 D(s)}{c_2} \times [c_3 - \log p_z(Z_s) - \sum_{k=1}^{K} \log |det J_{f_k}|]ds,$  (30)

where $c_1$ $c_2$ and $c_3$ are effect-free constants. To minimize $D_{KL}$, the Loss Function is refined as

$L_s = \sum_{s \sim p_s(s)} D(s)[-\log p_z(Z_s) - \sum_{k=1}^{K} \log |det J_{f_k}|].$ (31)

## IV. IMPLEMENTATION

According to methodological framework in Section III, the implementation procedure is presented in Fig.3 accordingly.

### A. Naturalistic Driving Data and Distribution Fitting

We collect naturalistic driving data from HighD [27], a high-quality dataset of naturalistic vehicle trajectories, maneuvers, and relevant background vehicles' information recorded on German highways.

Gaussian Mixed Model (GMM) is used to represent the joint distribution of scenario variables in the naturalistic driving environment. For a Gaussian Mixture Model with $K$ components, the $k^{th}$ component has a mean of $\vec{\mu}_k$, and covariance matrix of $\sum_k$. The mixture component weights are defined as $\Phi_k$ for component $C_k$, with the constraint that $\sum_{i=1}^{K} \Phi_i = 1$.

### B. The Network for Normalizing Flow

As is shown in (27) and (28), the NF consists of compositions of two types of prototypical transformations. In our model, we map one variable in (28) and keep others unchanged in (27) in one affine coupling layer. One variable requires 4 neural networks (forward and inverse), and 16 neural networks of 4 affine coupling layers will be trained jointly. Here, we apply two-layer linear neural networks. Hyperparameters of the networks are listed in Table 1.

TABLE I. HYPERPARAMETERS OF THE NETWORKS

| Hyperparameters | Neural network |
| --- | --- |
| Hidden1 | 512 |
| Activation Function 1 | LeakyReLU |
| Hidden2 | 512 |
| Activation Function 2 | Tanh |
| Combining coupling layers | 16 |
| Optimizer | 1e-3 |
| Learning rate | Adam |

Considering limited parameters in the simple testing scenario, $m$ and $s$ are input as vectors, which can be handled by a linear network. It is notable that the simple network for now can be improved to a sophisticated structure. When the testing scenario $X_T$ involves more background vehicles, $m$ and $s$ can be input as three-dimensional tensors, necessitating an upgrade of the network structure to a convolutional neural network such as Glow [28].

### C. Temporal Importance Sampling

For a scenario $X_T$, the values of variables at all timesteps cannot be sampled simultaneously. Instead, it follows a "stepwise sampling while stepwise simulating" process. Due to the complex and non-parametric probability density function of Importance Sampling distribution, the accept-reject sampling method is adopted. Sampling maneuvers $m_t$ from Importance Sampling distribution is as follows:

*Step 1*  Randomly sample the initial state $s_1$ from $q^*_s(s)$.

*Step 2*  Sample a candidate $m'_t$ for the next timestep from a uniform distribution:

$m'_t \sim U[m_{min}, m_{max}],$  (32)

where $m_{min}$ and $m_{max}$ are the minimum and maximum values of the maneuvers of background vehicles in the naturalistic driving distribution.

*Step 3*  Sample the candidate probability $p'_t$ of $m'_t$ from a uniform distribution:

$p'_t \sim U[0, p(m|s)_{max}].$  (33)

*Step 4*  According to the accept-reject sampling principle, accept the candidate $m'_t$ that satisfies (34) as the $m_t$,

$$p'_t \leq \frac{q^*_{m,s}(m'_t, s_t)}{q^*_s(s_t)} \quad (34)$$

*Step 5* Calculate $s_{t+1} = C(s_t, m_t)$

*Step 6* Repeat from *Step 2* until a risky event occurs or $t = T$.

The computational time complexity of this algorithm is influenced by $T$ and the sampling acceptance rate.

## V. EXPERIMENTS

### A. Vehicles in Simulated Scenarios

The SUT in our experiment is the rear-end collision avoidance functionality fulfilled by the Intelligent Driver Model (IDM, [29]). It involves an acceleration process in a free flow state and a deceleration process in congested flow. However, IDM can decelerate at a rate greater than the desired deceleration if the gap between vehicles becomes too narrow. This braking strategy makes IDM collision-free. In this study, we introduced an additional parameter $b_m = 4.5 \, m/s^2$ as a hard-coded cap on the deceleration to make IDM more realistic. The value of $b_m$ refers to international standards[30].

We considered a case involving only one AV. The scenarios variables included the speed of the AV, the speed of the leading vehicle, the distance between them, and the acceleration of the leading vehicle.

As a tentative practice, we only considered go-straight maneuvers of the leading vehicle in this experiment.

All variables were normalized for the efficiency of training. Naturalistic distribution of scenario variables fitted by GMM is shown in Fig.4.

### B. Performance Evaluaton of Accelerated Validation

The experiment was run on a machine with Intel Core I7-13700F CPU @ 2.10 GHz, 16 GB RAM and NVIDIA QUADRO RTX 4000 GPU.

The ratio of the training set to the test set is 8:2. Fig.5 shows the training loss of NF. The loss gradually descended and converged after 60 epochs. The loss of test set was 0.012, which verified the generalization ability of the trained model to some extent.

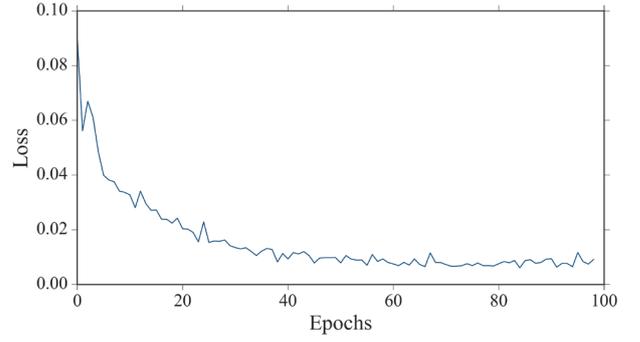

Fig. 5. Training loss of NF model

For each scenario, the state of Car-following was updated 10 steps forward according to SUT and background vehicles. Fig.6 shows the distribution shift of scenario variables. The red curves describe the scenarios generated from the risky event distribution learned by NF. With clear and intuitive differences, the distance between the two vehicles narrows down significantly, and the leading vehicle' acceleration decreases noticeably. As shown in Figure 7, after inverse normalization, we observe that the mean acceleration of the leading vehicle sampled from risky event distribution is less than 0, indicating that the risky events are primarily caused by the continuous deceleration of the leading vehicle. In contrast, the acceleration sampled from the naturalistic driving distribution is more evenly distributed around 0.

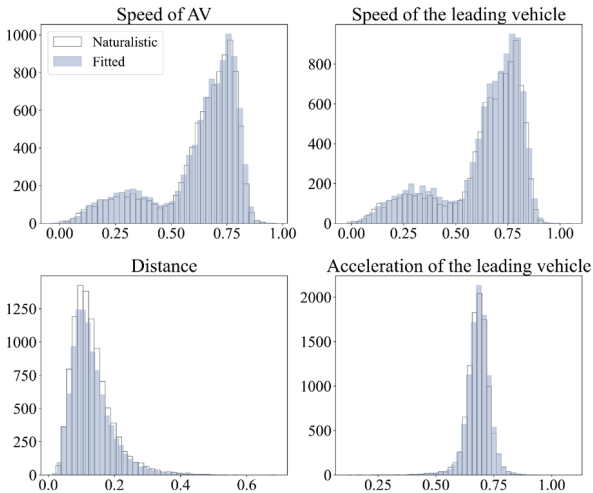

Fig. 4. Fitted naturalistic distribution of scenario variables

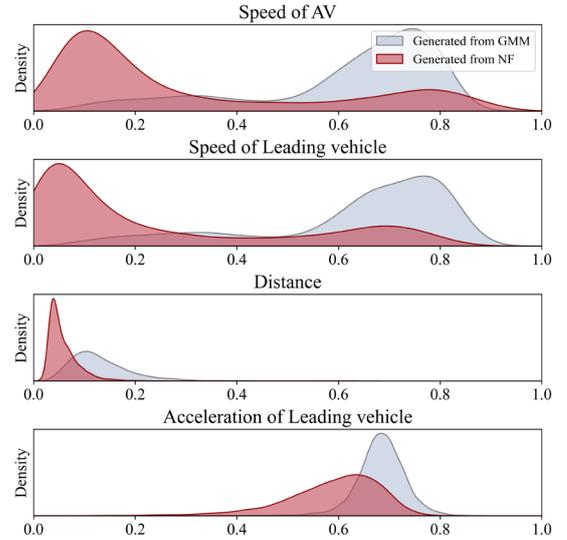

Fig. 6. Distribution shift of scenarios variables

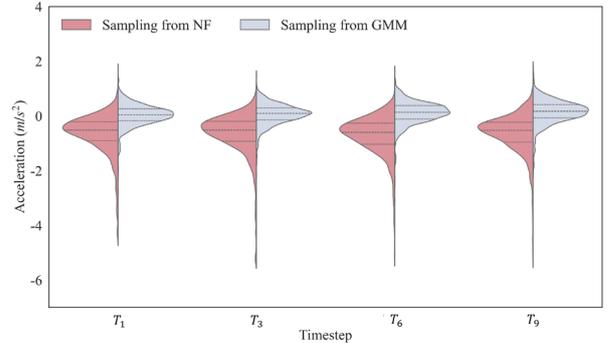

Fig. 7. Acceleration of the leading vehicle

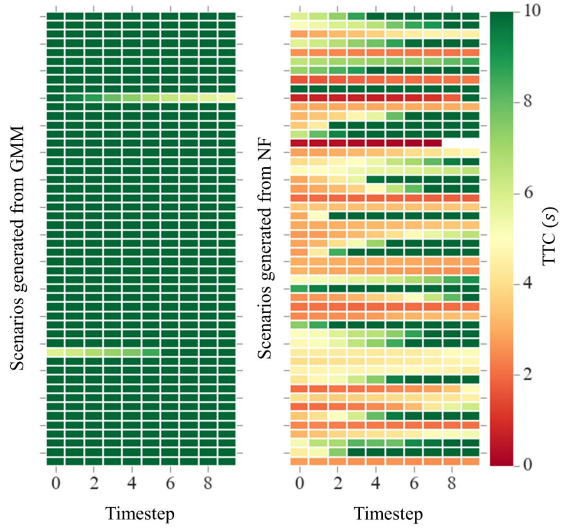

Fig. 8. The Temporal TTC of testing scenarios

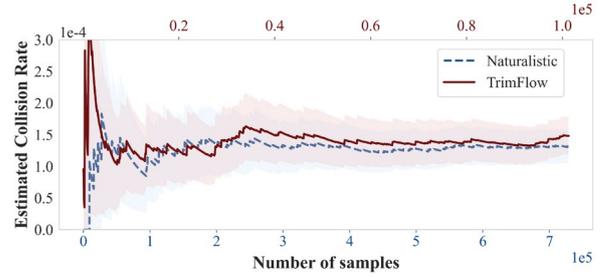

Fig. 9. Comparison of the collision rate estimation.

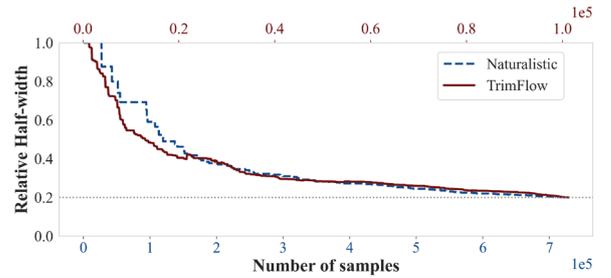

Fig. 10. Comparison of the decrease of relative half width.

Among scenarios generated by NF, 97.5% of them have a minimum TTC less than 10s, whereas this proportion is only 4.4% in the testing scenarios generated by GMM. The temporal TTC of 50 randomly sampled testing scenarios from GMM and NF are separately visualized in Fig.8. It can be concluded that although the SUT is well-equipped with rear-end collision avoidance functionality, NF indeed learnt the distribution of risky events and provided more hazardous test scenarios.

The performance of TrimFlow in Car-following scenarios is shown in Fig.9 and Fig.10. If sampling from the naturalistic driving distribution, the estimated collision rate converged to $1.33 \times 10^{-4}$. The relative half width reached the required accuracy (relative half width $\omega < 0.2$) after $7.2 \times 10^5$ scenario tests. Based on the TrimFlow method, the estimated collision rate converged to $1.48 \times 10^{-4}$. The minor gap between the two estimated collision rates was believed to be acceptable because of the stochastic sampling. The relative half width reached the required accuracy after $1.0 \times 10^5$ scenario tests, which meant that TrimFlow reduced the number of tests by 86.1% compared to tests from naturalistic driving distribution.

Consideration of temporal testing scenarios with continuous maneuver variables brings great difficulties to validation methods. Compared to [8, 14], since TrimFlow doesn't need known risky rare samples in its training set as the warm start for distribution learning, the effectiveness can be viewed as an outcome that holds potential for further improvement. We believe that the main reason is that the network structure of the NF is too simplistic and the optimal $q_s^*(s)$, $q_{m,s}^*(m,s)$ have not been well learnt. Although we can successfully sample much more collision scenarios from NF, their corresponding probabilities are not amplified adequately. It can also be refined by iterative training sets as was adopted in [8], where the simulated collision scenarios can reinforce the training set and enhance the distribution learning.

## VI. CONCLUSION

In this paper, we devise the TrimFlow method to represent, generate, and reweight the distribution of risky rare events, which contributes to the accelerated validation for AVs. We apply it on Car-following scenarios and verify its effectiveness. The test results indicate that, to obtain the reliable statistical estimation of collision rate, the proposed method can effectively save the required tests. Considering the tests reduction can be further broken through, different formations of NFs and other feasible generative models will be on trial.

Developing accelerated validation methods is vital for the future of AVs. In our subsequent work, more advanced SUT, such as Hardware-in-the-Loop and Vehicle-in-the-Loop will be used to explore the capability of TrimFlow.